# Confederated learning in healthcare: training machine learning models using disconnected data separated by individual, data type and identity for Large-Scale Health System Intelligence

Dianbo Liu[1,2,3,4,], Kathe Fox[3,5], Griffin Weber[3], Tim Miller[1,2]
1.Computational Health Informatics Program, Boston Children's Hospital, Boston, MA
2.Department of Pediatrics, Harvard Medical School, Boston, MA
3.Department of Biomedical Informatics, Harvard Medical School, Boston, MA
4. Computer Science & Artificial Intelligence Laboratory, MIT, Cambridge, MA
5. Aetna, CVS Health, Boston, MA
Corresponds to: dianbo@mit.edu / liudianbo@gmail.com

## Abstract

A patient's health information is generally fragmented across silos because it follows how care is delivered: multiple providers in multiple settings. Though it is technically feasible to reunite data for analysis in a manner that underpins a rapid learning healthcare system, privacy concerns and regulatory barriers limit data centralization for this purpose. Machine learning can be conducted in a federated manner on patient datasets with the same set of variables, but separated across storage. But federated learning cannot handle the situation where different data types for a given patient are separated vertically across different organizations and when patient ID matching across different institutions is difficult. We call methods that enable machine learning model training on data separated by two or more dimensions "confederated machine learning." We propose and evaluate confederated learning for training machine learning models to stratify the risk of several diseases among silos when data are horizontally separated by individual, vertically separated by data type, and separated by identity without patient ID matching. The confederated learning method can be intuitively understood as a distributed learning method with representation learning, generative model, imputation method and data augmentation elements.

## Introduction

**Significance.** Access to a large amount of high quality data is possibly the most important factor for success in advancing medicine with machine learning. However, valuable healthcare data are usually distributed across isolated silos, and there are complex operational and regulatory concerns. Data on patient populations are often *horizontally* separated, *by individual*, from each other across different practices and health systems. In addition, individual patient data are often vertically separated, *by data type*, across the sites of care, service, and testing. Furthemore, it is often not possible to match patient IDs across different silos in the healthcare system due to operational and privacy issues, which separates patients' data *by identity*. Traditionally, federated machine learning refers to distributed learning on horizontally separated data (Yue Zhao, Meng Li, Liangzhen Lai, Naveen Suda, Damon Civin, Vikas Chandra 2018; Cano, Ignacio, Markus Weimer, Dhruv Mahajan, Carlo Curino, and Giovanni Matteo Fumarola 2016; H. Brendan McMahan, Eider Moore, Daniel Ramage, Seth Hampson, Blaise Agüera y Arcas 2016). Algorithms are sent to different data silos (sometimes called data nodes) for training. Models obtained are aggregated for inference. Federated learning can reduce data duplication and costs associated with data transfer, while increasing security and shoring up institutional autonomy (Geyer, R. C., Klein, T., & Nabi, M. 2017; al. 2016) .

Notably, a patient's vertically separated data may span data types--for example, diagnostic, pharmacy, laboratory, and social services. Machine learning on vertically separated data has used a split neuron network (Praneeth et al. 2018) and homomorphic encryption (Praneeth et al. 2018; Stephen et al. 2017). However, these new methods require patient ID matching, information communication at each computational cycle and state-of-art computational resource organization, which are usually impractical in many healthcare systems where support for data analysis is not the first priority, high speed synchronized computation resources are often not available, and data availability is inconsistent.

To accelerate a scalable and collaborative rapid learning health system (Friedman, Wong, and Blumenthal 2010; Mandl et al. 2014), we propose a confederated machine learning method that trains machine learning models on data separated by individual and data type, using a 3-step approach from data distributed across silos (Qi, Huang, and Peng 2017; Zhang and Xiao 2015; Zhai, Peng, and Xiao 2014). Our confederated learning method does not require individual ID matching, frequent information exchange at each training epoch nor state-of-the-art distributed computing infrastructures. As such, it should be readily implementable, using existing health information infrastructure.

Our proposed model assumes the existence of one central institution with limited amounts of complete and fully connected data including all data types. This kind of centralization is the exception in American medicine but may be present in some large institutions (e.g., academic medical centers, integrated care organizations). While such institutions may be able to train some of their own machine learning models, for many learning tasks they still have much to gain by participating in confederated learning networks, as our experiments show. When more accurate and generalizable models are needed, data across all silos in the network need to be included in the training process in a federated manner. Our method enables efficient utilization of data from all participating silos in a federated data network. Examples of such federated medical data networks include i2b2 network(Murphy et al. 2010), Undiagnosed Diseases Network (UDN)(Gahl, Wise, and Ashley 2015) and European Reference Networks(Gahl, Wise, and Ashley 2015; Héon-Klin 2017). In this article, we conducted simulated experiments using real world data to better understand how confederated machine learning could enable training of models using medical data available from federated networks. To illustrate how confederated learning works, diagnosis, medication fills and lab tests data from a major commercial healthcare insurer in the U.S. were used to simulate a federated medical data network consisting of many clinics with disease and diagnosis information, pharmacies with medication fill records and clinical laboratories with lab test records. To mimic a real use case, we assume that only the central analyzer institution has direct access to aggregated data of all three types from a single arbitrarily picked state in the U.S. The central analyzer institution is able to send and receive machine learning models to and from other silos in the network but does not have direct access to other silos' data. Using this simulated system, we conducted proof-of-concept experiments to show how this confederated learning method makes it possible to train models on data with both dimensions of separation mentioned above in a large federated medical data network.

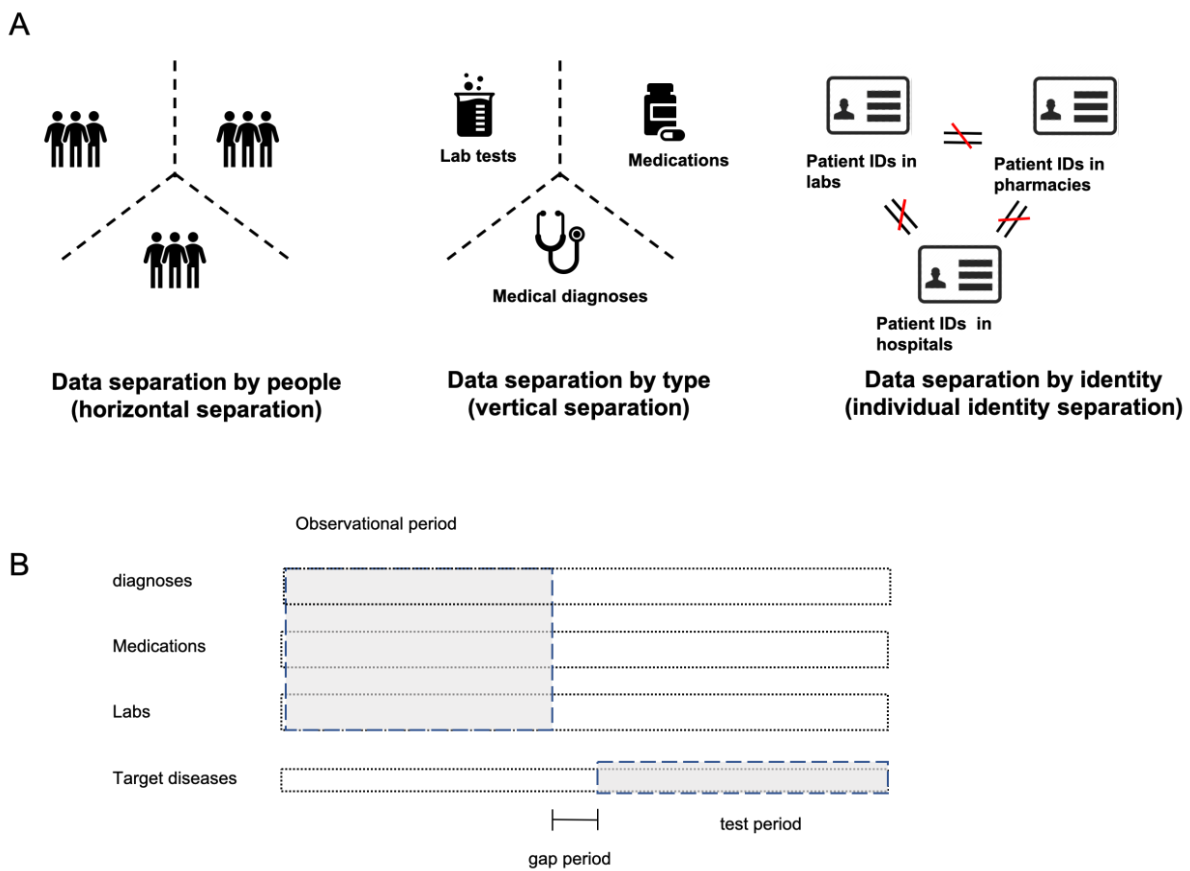

**Figure 1. (A) Three degrees of separation.** *Separation by individual* or *horizontal separation* refers to fragmentation of an individual's data across silos, for example across hospitals and clinics. *Separation by data type* or *vertical separation* refers to differences in the domain, semantics and structure of the data, for example, data from pharmacies, clinics and labs, each in their own nodes. *Separation by identity* refers to the problem of not being able to match individuals using their IDs, mostly due to inconsistency IDs among silos. **(B) Study period.** Patient's data are divided into three periods. Observational period is 24 months, gap period is 1 week and follow-up period is 23 months and 3 weeks. Diagnoses, medications and lab tests data of each patient in the observational period were used as predictive features for target diseases, such as diabetes in the test period. The 1-week gap period is introduced to avoid complications of encounters happening directly before diagnosis of target diseases.

The confederated learning method we developed consists of three steps: Step 1) Conditional generative adversarial networks with matching loss (cGAN) were trained using data from the central analyzer to infer one data type from another, for example, inferring medications using diagnoses. Generative (cGAN) models were used in this study because a considerable percentage of individuals has not paired data types. For instance, a patient may only have his or her diagnoses in the database but not medication information due to insurance enrolment. cGAN can utilize data with paired information by minimizing matching loss and data without paired information by minimizing adversarial loss. Step 2) Missing data types from each silo were inferred using the model trained in step 1. Step 3) Task-specific models, such as a model to predict diagnoses of diabetes, were trained in a federated manner across all silos simultaneously.

## Results

We conducted experiments to train disease prediction models using confederated learning on a large nationwide health insurance dataset from the U.S that is split into 99 silos. The models stratify individuals by their risk of diabetes, psychological disorders or ischemic heart disease in the next two years, using diagnoses, medication claims and clinical lab test records of patients (See Methods section for details). The goal of these experiments is to test whether a confederated learning approach can _simultaneously_ address the two types of separation mentioned above.

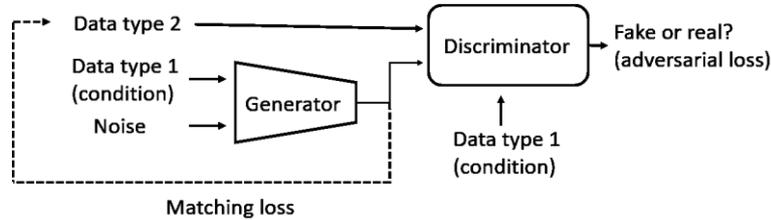

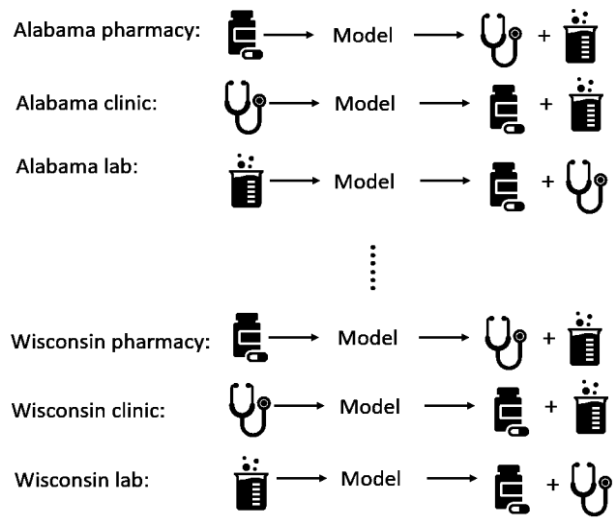

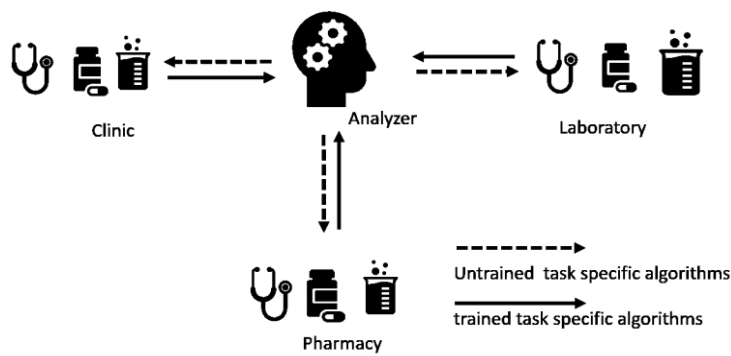

**Figure 2. Confederated learning trains machine learning models using disconnected data separated by individual, data type and identity for Large-Scale Health System Intelligence.** The method consists of 3 steps: Step 1) Conditional generative adversarial networks with both adversarial and matching loss were trained using data from the central analyzer to infer one data type from another, eg. inferring medications using diagnoses (Mehdi Mirza 2014; Zhu et al. 2017). Step 2) Missing data types from each silo were inferred using the model trained in step 1. Step 3) Task-specific models , such as a model to

predict diagnoses of diabetes, were trained in a federated manner across all silos simultaneously.

Table 1. Number of individuals and patients in each silo

| Place | Total number of people | Diabetes | Psychological disorders | Ischemic heart disease |
|---|---|---|---|---|
| Alabama | 154 | 23 | 18 | 24 |
| Arizona | 485 | 89 | 50 | 44 |
| Arkansas | 163 | 27 | 21 | 19 |
| California | 9074 | 744 | 232 | 242 |
| Colorado | 326 | 37 | 36 | 14 |
| Delaware | 1979 | 481 | 205 | 210 |
| District of Columbia | 254 | 66 | 26 | 22 |
| Florida | 4759 | 1096 | 503 | 564 |
| Georgia | 2279 | 536 | 258 | 230 |
| Illinois | 1522 | 335 | 159 | 151 |
| Indiana | 888 | 181 | 102 | 97 |
| Kansas | 124 | 31 | 9 | 8 |
| Kentucky | 641 | 126 | 81 | 77 |
| Louisiana | 399 | 76 | 47 | 42 |
| Maryland | 1889 | 508 | 223 | 192 |
| Michigan | 2890 | 565 | 301 | 312 |
| Minnesota | 163 | 36 | 9 | 18 |
| Mississippi | 233 | 45 | 37 | 30 |
| Missouri | 229 | 43 | 21 | 24 |
| Nevada | 1898 | 524 | 193 | 195 |
| New York | 8188 | 1846 | 763 | 991 |
| North Carolina | 1260 | 264 | 162 | 116 |
| Ohio | 7346 | 1497 | 974 | 689 |
| Oklahoma | 512 | 87 | 67 | 62 |
| Oregon | 134 | 18 | 15 | 7 |
| Pennsylvania | 16557 | 3562 | 1794 | 1733 |
| South Carolina | 839 | 170 | 100 | 87 |
| Tennessee | 1439 | 291 | 164 | 176 |

| Texas | 11411 | 2571 | 1259 | 1330 |
| --- | --- | --- | --- | --- |
| Utah | 114 | 20 | 11 | 5 |
| Virginia | 1905 | 514 | 170 | 134 |
| Washington | 514 | 81 | 65 | 45 |
| West Virginia | 1391 | 287 | 165 | 135 |
| Wisconsin | 184 | 47 | 25 | 19 |

**Disease-predicting models can be trained efficiently using confederated learning on disconnected data.** The confederated learning approach achieved an AUCROC of 0.79, AUCPR of 0.47, PPV of 0.56 and NPV of 0.81 predicting onset of diabetes in the test period, AUCROC of 0.72, AUCPR of 0.24, PPV of 0.36 and NPV of 0.91 predicting psychological disorders and AUCROC of 0.72, AUCPR of 0.24, PPV of 0.36 and NPV of 0.91 in predicting psychological disorders (Table 2).

In order to understand whether the confederated learning method efficiently utilized data separated by all three levels, we conducted 3 control experiments. First, to create an effective upper bound on confederated learning, we aggregated all the data from all silos such that data are not disconnected at all, and the models are trained in a standard centralized manner (Table 2). As expected, this configuration is superior to federated or confederated models. Second, when models were only trained on data from the central analyzer, the performance is inferior to models trained in a confederated manner. Lastly, when we conducted federated learning across all silos with only a single data type (e.g., only diagnoses), the models obtained also showed poorer results in all three tasks compared with confederated learning.

**Confederated learning using different states as the central analyzer.** In the previous sections, we randomly chose one state to represent the "central analyzer" (California), with access to fully connected data from clinics, pharmacies and labs. In order to understand the sensitivity of performance to the characteristics of the central analyzer, we conducted experiments to compare mean AUCROC and AUCPR across all targeted diseases and conditions using data from each of the 34 states as the central analyzer (Table 3). To understand the relation between confederated model performance and number of people with the central analyzer state, we plot mean AUCROC vs. number of people in central analyzer (Figure 3 A) and increase in mean AUCROC of confederated learning model compared with a model trained with only data from the central analyzer (Figure 3 B). It is observed that, as expected, the more patients the center analyzer has, the better performance the confederated model will be. This is intuitively understandable as the imputation model will be better trained with a larger number of patients at the center analyzer. In addition, to understand potential incentives for different sites to join a federated healthcare network, we also include a model trained using data from each state to compare performance. Confederated models outperformed models trained using only data from the center analyzer states in 32 out of 34 states by mean AUCROC and all 34 states by AUCPR. The performance was measured in a testing set sampled from across the country. This suggested that even sites with large numbers of patients would benefit from participating in confederated learning to increase their model performance. In addition, magnitudes of increase in absolute value of mean AUCROC grow with the number of people in the central analyzers and saturated when the central analyzer has around 5000 people. Similar trends were observed for AUCPR. These observations indicate that, at least in the study setting and tasks in this study, central analyzers with larger sample size will help improve performance of confederated models trained in the system and even very large silos would benefit from confederated learning and, therefore, should consider joining the network.

**Table 2.** Performance of disease predicting models trained using confederated learning on medical data distributed in 99 silos separated by individual, data type and identity.

|  | AUCROC | AUCPR | PPV | NPV |
|---|---|---|---|---|
| **Diabetes** | | | | |
| Data with no separation (centralized) | 0.824 | 0.526 | 0.620 | 0.809 |
| Use only data with central analyzer | 0.774 | 0.451 | 0.554 | 0.808 |
| Federated learning using diagnosis* | 0.775 | 0.465 | 0.614 | 0.809 |
| Confederated learning | 0.787 | 0.472 | 0.563 | 0.809 |
| **Psychological disorders** | | | | |
| Data with no separation (centralized) | 0.757 | 0.266 | 0.340 | 0.905 |
| Use only data with central analyzer | 0.647 | 0.163 | 0.230 | 0.903 |
| Federated learning using diagnosis* | 0.590 | 0.126 | 0.134 | 0.900 |
| Confederated learning | 0.718 | 0.239 | 0.356 | 0.909 |
| **Ischemic heart disease** | | | | |
| Data with no separation (centralized) | 0.721 | 0.185 | 0.201 | 0.907 |
| Use only data with central analyzer | 0.679 | 0.160 | 0.197 | 0.907 |
| Federated learning using diagnosis* | 0.657 | 0.151 | 0.176 | 0.907 |
| Confederated learning | 0.698 | 0.169 | 0.199 | 0.909 |

*As models obtained from federated learning on diagnosis had better performance than models trained on medications or lab tests, only results of models trained on diagnoses are shown in this table.

**Table 3**. Mean AUCROC and AUCPR across different tasks of confederated model trained different states as the central analyzers or model trained using data from the central analyzers.

| Center state | Total number of people | Confederated learning AUCROC (mean/sd) | Confederated learning AUCPR (mean/sd) | Single state AUCROC (mean/sd) | Single state AUCPR (mean/sd) |
|---|---|---|---|---|---|
| Alabama | 154 | 0.544(0.010) | 0.153(0.080) | 0.546(0.048) | 0.131(0.036) |
| Arizona | 485 | 0.564(0.077) | 0.183(0.136) | 0.522(0.005) | 0.144(0.073) |
| Arkansas | 163 | 0.564(0.075) | 0.182(0.143) | 0.559(0.045) | 0.116(0.040) |
| Colorado | 326 | 0.549(0.036) | 0.145(0.095) | 0.576(0.085) | 0.109(0.029) |
| Delaware | 1979 | 0.688(0.089) | 0.269(0.170) | 0.533(0.022) | 0.137(0.081) |
| District of Columbia | 254 | 0.567(0.088) | 0.189(0.145) | 0.575(0.055) | 0.112(0.038) |
| Florida | 4759 | 0.723(0.054) | 0.276(0.149) | 0.553(0.009) | 0.157(0.069) |
| Georgia | 2279 | 0.696(0.049) | 0.267(0.148) | 0.532(0.025) | 0.138(0.085) |
| Illinois | 1522 | 0.680(0.064) | 0.261(0.138) | 0.528(0.009) | 0.127(0.061) |
| Indiana | 888 | 0.647(0.078) | 0.235(0.144) | 0.524(0.031) | 0.131(0.069) |
| Kansas | 124 | 0.576(0.049) | 0.164(0.135) | 0.562(0.050) | 0.117(0.037) |
| Kentucky | 641 | 0.580(0.095) | 0.204(0.157) | 0.539(0.014) | 0.139(0.057) |
| Louisiana | 399 | 0.554(0.074) | 0.192(0.153) | 0.557(0.039) | 0.114(0.047) |
| Maryland | 1889 | 0.667(0.053) | 0.246(0.137) | 0.528(0.015) | 0.135(0.057) |
| Michigan | 2890 | 0.702(0.052) | 0.278(0.149) | 0.532(0.058) | 0.164(0.108) |
| **Minnesota** | 163 | 0.559(0.068) | 0.174(0.136) | 0.528(0.035) | 0.124(0.047) |
| **Mississippi** | 233 | 0.553(0.067) | 0.181(0.121) | 0.546(0.023) | 0.126(0.043) |
| **Missouri** | 229 | 0.569(0.056) | 0.168(0.132) | 0.563(0.048) | 0.115(0.044) |
| **Nevada** | 1898 | 0.658(0.036) | 0.240(0.133) | 0.509(0.004) | 0.141(0.074) |
| **New York** | 8188 | 0.737(0.037) | 0.301(0.161) | 0.594(0.061) | 0.192(0.122) |
| **North Carolina** | 1260 | 0.650(0.059) | 0.241(0.147) | 0.559(0.063) | 0.114(0.042) |
| **Ohio** | 7346 | 0.737(0.051) | 0.292(0.149) | 0.585(0.050) | 0.186(0.113) |
| **Oklahoma** | 512 | 0.587(0.061) | 0.207(0.134) | 0.529(0.014) | 0.128(0.052) |
| **Oregon** | 134 | 0.550(0.043) | 0.151(0.104) | 0.532(0.036) | 0.125(0.043) |
| **Pennsylvania** | 16557 | 0.753(0.045) | 0.305(0.153) | 0.610(0.119) | 0.214(0.171) |
| **South Carolina** | 839 | 0.595(0.094) | 0.218(0.147) | 0.527(0.019) | 0.147(0.079) |
| **Tennessee** | 1439 | 0.673(0.039) | 0.250(0.141) | 0.531(0.008) | 0.142(0.073) |
| **Texas** | 11411 | 0.744(0.047) | 0.298(0.162) | 0.586(0.116) | 0.207(0.153) |
| **Utah** | 114 | 0.542(0.037) | 0.154(0.101) | 0.546(0.061) | 0.117(0.038) |
| **Virginia** | 1905 | 0.662(0.060) | 0.244(0.122) | 0.532(0.028) | 0.129(0.072) |
| **Washington** | 514 | 0.531(0.044) | 0.170(0.103) | 0.532(0.033) | 0.123(0.049) |
| **West Virginia** | 1391 | 0.652(0.058) | 0.242(0.139) | 0.525(0.022) | 0.132(0.068) |
| **Wisconsin** | 184 | 0.559(0.059) | 0.166(0.125) | 0.552(0.052) | 0.115(0.043) |
| **Average** | 2416(3779) | 0.627(0.072) | 0.220(0.050) | 0.549(0.026) | 0.139(0.028) |

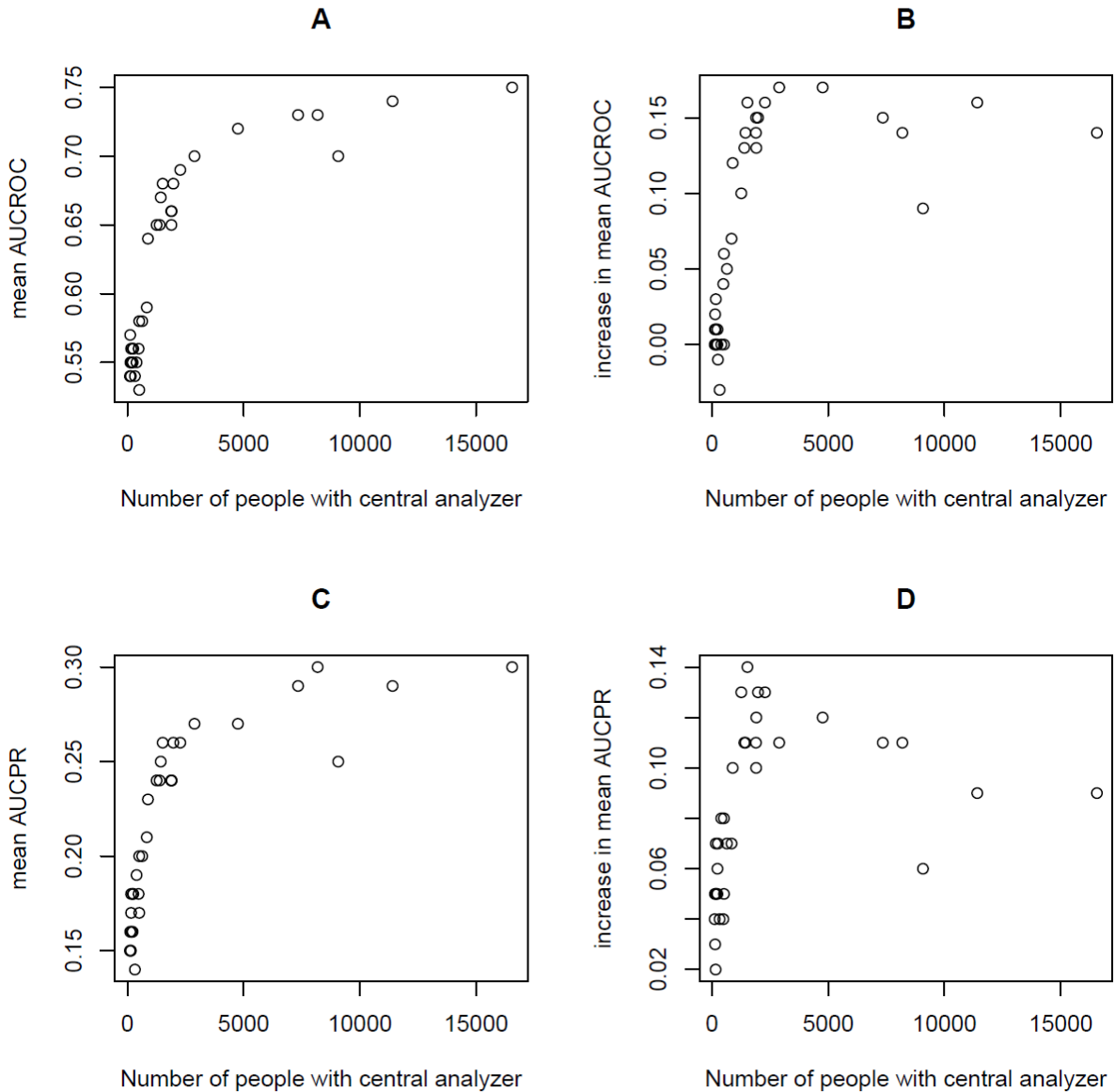

Figure 3. (A) Mean AUCROC of confederated learning models vs. number of people with fully connected data available in central analyzer. Each point represents a confederated experiment conducted with a specific state, eg. California, as the central analyzer. Each state has a different number of people in the database. (B) Improvement of mean AUCROC when using confederated learning models vs. number of people with fully connected data in the central analyzer. Models trained only using data from the central analyzer was used as the baseline. (C) Mean AUCPR of confederated learning models vs. number of people with fully connected data in central analyzer. (D) Improvement of mean AUCPR when using confederated learning models vs. number of people with fully connected data in the central analyzer. Models trained only using data from the central analyzer was used as the baseline.

## Methods

**Data source and cohort.** The study uses claims data from a major U.S. health plan. Elements include the insurance plan type and coverage periods, age, sex, medications, and diagnoses associated with billed medical services, from July 1 2016 to June 31 2019. The dataset contains an indicator for insurance coverage by month. Only beneficiaries having full medical and pharmacy insurance coverage during the 48-month period were included. The study period is divided into a 12-month observational period, a 1-week gap period and a test period of 23 months and 3 weeks (Figure 1B). Individuals above age of 65 not enrolled in the Medicare Advantage program were excluded to ensure completeness of the private and public insurance data. A total of 82,143 individuals were included in this study.

The input features to the confederated machine learning model include diagnoses as ICD 10 codes, medications represented as National Drug Codes (NDC) and lab tests (encoded as LOINC codes). Lab test results were not available for this study. On average, each individual has 13.6 diagnoses, 6.9 prescriptions, and 7.4 LOINC codes during the 24 month observational period.

**Study outcome.** Diagnoses in the claim data were originally provided as an online International Classification of Diseases, Tenth Revision (ICD-10). ICD10 codes of target diseases were selected according to mapping between ICD 10 codes and Phecodes from PheWASCatalog (https://phewascatalog.org/). Phecodes were defined by hierarchical grouping of ICD codes and were originally used for phenome-wide associations studies(Bastarache et al. 2018). For each member, we marked with a binary outcome variable (0 or 1) whether a person had any claims related to each of the target diseases during the follow-up period. ICD-10 codes corresponding to PheCode 249-250 were used to define diabetes. ICD-10 codes corresponding to PheCode 295-306 were used to define psychological disorders and ICD-10 codes corresponding to PheCode 410-414.99 were used to define ischemic heart disease. Among the individuals, 16,824 had diabetes, 8,265 had psychological disorders and 8,044 had ischemic heart disease in the test periods.

**Study setting.** Data from 34 states in the U.S. were included into this study. The central analyzer has access to all three data types and ability to match patients' IDs across types from one state. In this study, data from California with information of 5433 individuals was used as the central analyzer's dataset. Data from the remaining 33 states were divided into 99 silos by state and data type. We assume individual ID matching among silos was not possible, which is common in healthcare. For a specific silos $s \in \{1,2,\ldots S\}$ with $S = 99$ in this study, Each individual $i$ has either a diagnosis vector $X_{si}^{diag}$ or a medication claim vector $X_{si}^{med}$ from pharmacy, or lab test vector $X_{si}^{lab}$ from clinical lab, where in each silo $i \in \{1,2,\ldots,n^s\}$ with $n^s$ being the number of beneficiaries in the silo. Our confederated learning method aims to train disease classification models using disconnected data from all the silos.

Claims for diagnoses, medications and lab tests during the observation period are the input features. The output of the classifier is a binary variable indicating whether the beneficiary had a target disease during the test period. Diabetes, psychological disorders and ischemic heart disease were chosen as targeted diseases in the test period due to their clinical importance. We simulated horizontal separation by separating the data for beneficiaries by U.S. state of residence. We simulated vertical separation by assuming that beneficiaries' diagnoses are only available in clinics, medication claims data are only kept in pharmacies and lab data only in labs. We simulated separation by identity by assuming individual ID matching was not possible among silos. Data is presumed to not be shared among different organizations nor across state lines. In total, besides the central analyzer, we simulated data distributed across 99 distinct silos including 33 clinical silos, 33 pharmacy silos and 33 lab silos.

**Confederated learning**. In *step 1*, using available data from the central analyzer, including diagnoses ($X_{si}^{diag}$), medications ($X_{si}^{med}$) and lab tests ($X_{si}^{lab}$), a conditional generative adversarial network (cGAN) was trained for each pair of data types (Zhu et al. 2017; Mehdi Mirza 2014).. For example, inferring medications using diagnoses. To train the generator, two losses were used. A least square adversarial loss was included to minimize errors on discriminator. A L1 matching loss between the generated data and real data was included that encourages the outputs to be close to observed data(Isola et al. 2017; Mao et al. 2019). A random Gaussian noise vector with length 100 was used as the source of stochasticity. In addition, another classifier was trained on the central analyzer to map each data type to the target disease label $Y_{s,i}^{disease}$, which will be used in step 2. There are two reasons why we used a cGAN based method instead of a deterministic supervised method to infer missing data type. First, some of the individuals do not have all data types available in the data set; cGAN based methods are able to utilize these incomplete data for discriminator training. Secondly, we are more interested in the potential distribution of a data type rather than a point estimate due to the heterogeneous nature of healthcare.

In *Step 2*, the cGANs and target disease classifiers obtained from step 1 were passed to each of the 99 silos. In each silo, $X_{s,i}^{diag}$, $X_{s,i}^{med}$ and $X_{s,i}^{lab}$, if not available in the corresponding silo, were inferred using cGAN based on data type available (See Methods for details). The binary labels $Y_{s,i}^{disease}$ of targeted diseases were inferred from $X$ using the classifier model obtained from step 1.

In *step 3*, after the data inference step, each silo has all three types of data, namely, diagnoses, medication and lab tests, as well as the inferred binary labels of targeted diseases. Therefore, a traditional federated learning method can be applied. We applied federated averaging algorithms for step 3 due to its popularity and robustness (H. Brendan McMahan, Eider Moore, Daniel Ramage, Seth Hampson, Blaise Agüera y Arcas 2016).

Multi-layer neural network models with batch normalization and drop out were used for both generators and discriminators in the cGANs. Leaky ReLU was used as an activation function for hidden layers. A separate cGAN was built for each pair of data types. Another multi-layer neural network was used as classifiers to map each data type to binary labels. One classifier was built for each of the three disease labels.

**Model, training and performance evaluation details.** In step 3, the goal of the training was to minimize the binary cross entropy, a metric for binary classification error, without moving any data out of their data silos. The objective function to minimize for classification model $f(X^{diag}, X^{med}, X^{lab}, \Theta)$ is:

$$L(X^{diag}, X^{med}, X^{lab}, \Theta) = \sum_{s=1}^{S} \sum_{i=1}^{n_i} -(Y_{s,i} \log(f(X^{diag}, X^{med}, X^{lab}, \Theta))) + (1 - Y_{s,i}) \log(1 - f(X^{diag}, X^{med}, X^{lab}, \Theta)))$$

Where $\Theta$ is the parameter of model $f$.

As data were not allowed to be moved out from their silos, it is not possible to train $f$ by minimizing $L(X^{diag}, X^{med}, X^{lab}, \Theta)$ in a centralized manner. Therefore, we randomly initialized the parameters $\Theta$ as $\Theta_0$ and sent model $f$ and parameters $\Theta_0$ to pharmacies or clinical labs in each silo $s \in S$. In each of the 99 silos, two of $X^{diag}$, $X^{med}$ and $X^{lab}$ are inferred from the third data type.

In the pharmacy silos, the loss function is then calculated as:

$$L(\hat{X}^{diag}, X^{med}, \hat{X}^{lab}, \Theta_{s,t}) = \sum_{i=1}^{n_i} -(Y_{s,i} \log(f(\hat{X}^{diag}, X^{med}, \hat{X}^{lab}, \Theta_{s,t}))) + (1 - Y_{s,i}) \log(1 - f(\hat{X}^{diag}, X^{med}, \hat{X}^{lab}, \Theta_{s,t})))$$

Using stochastic gradient descent to minimize the loss, new parameters $\Theta_{s,t}$ were obtained. $t \in \{1,2,\ldots T\}$ stands for number of global loops (Algorithm 1). $n_s$ is the number of individuals in the corresponding silo. Hats in $\hat{X}^{diag}$ and $\hat{X}^{lab}$ indicates that the diagnoses data and lab test data were inferred from medication data.

Similarly, In the lab silos: $L(\hat{X}^{diag}, \hat{X}^{med}, X^{lab}, \Theta_{s,t}) = \sum_{i=1}^{n_s} -(Y_{s,i} \log(f(\hat{X}^{diag}, \hat{X}^{med}, X^{lab}, \Theta_{s,t}))) + (1 - Y_{s,i}) \log(1 - f(\hat{X}^{diag}, \hat{X}^{med}, X^{lab}, \Theta_{s,t})))$

In the clinic silos: $L(X^{diag}, \hat{X}^{med}, \hat{X}^{lab}, \Theta_{s,t}) = \sum_{i=1}^{n_s} -(Y_{s,i} \log(f(X^{diag}, \hat{X}^{med}, \hat{X}^{lab}, \Theta_{s,t}))) + (1 - Y_{s,i}) \log(1 - f(X^{diag}, \hat{X}^{med}, \hat{X}^{lab}, \Theta_{s,t})))$

After $\Theta_{s,t}$ were trained locally in each single silos, they were sent back to the analyzer for aggregation by weighted averaging: $\Theta_{t+1} = \frac{1}{S} \sum_{s=1}^{S} \frac{n_s}{N} \Theta_{s,t}$ where $N$ is the total number of beneficiaries included in the study from all states. $\Theta_{t+1}$ is then sent back to each silo to repeat the whole global cycle to obtain $\Theta_{t+2}$. The global training cycles were stopped when the loss of the predictive model on the validation set did not decrease for 3 consecutive cycles.

20% of randomly chosen beneficiaries from all silos were reserved as test data, and not included in the training set, and 20% individuals from the central analyzer were chosen as validation set to adjust hyperparameters and the rest were used as the training set. When conducting federated or confederated learning, data from 20% of the beneficiaries at each node were used as an internal validation set. After hyperparameter tuning, both the training set and validation set were used to train the model to test performance.

Performance evaluation included area under the receiver operating characteristic curve (AUCROC) and area under the precision recall curve (AUCPR), AUCPR was used because the data are imbalanced--there are many more people without targeted diseases than those with the diseases. Instead of following the common practice of choosing a threshold that sets the false positive rates to be equal to the false negative rate (equal error rate), we chose the threshold which is 95% quantile of the predicted score in the test set. We sought to favor a screening strategy and are willing to tolerate some false positives. Using this threshold, the positive predictive value (PPV) and negative predictive value (NPV), which are commonly used metrics in clinical settings, were calculated (Table 2) and used as performance metrics in addition to AUCROC and AUCPR.

**Data Availability**: Due to privacy concerns, the data are not publicly available.

## Discussion

In this study, we demonstrated that health data distributed across silos separated by individual and data type can be used to train machine learning models without moving or aggregating data. Our method obtains predictive accuracy competitive to a centralized upper bound in predicting risks of diabetes, psychological disorders or ischemic heart disease using previous diagnoses, medications and lab tests as inputs. We compared the performance of a confederated learning approach with models trained on centralized data, only data with the central analyzer or a single data type across silos. The experimental results suggested that confederated learning trained predictive models efficiently across disconnected silos.

To the best of our knowledge, confederated learning proposed in this study, is the very first method developed to train machine learning models on data with the three degrees of separation in healthcare settings. Compared with other methods for model training on horizontally and vertically separated data, this confederated learning algorithm does not require sophisticated computational infrastructure, such as homomorphic encryption, nor frequent gradient exchange. In addition, due to the nature of the algorithm, the confederated training works fine even if a whole data type, such as medication, is missing or not available for some patients. More importantly, patient matching or patient ID sharing is not required in our confederated setting, which are both difficult in many medical settings. Furthermore, using our simulated experiment, we show that even silos with large numbers of patients would benefit from participating in confederated learning.

We are pleased to see that disconnected data can be used in a confederated manner for model training. It is worth pointing out inter-data type correlation such as associations of medication orders with diagnoses have long been known in the medical community (Hu et al. 2017). Therefore, it is intuitively understandable why the inter-data type imputation works in medical settings presented in this work. However, one interesting implication is that the method essentially multiplies out the number of patients in the data set, by taking each data type for a given patient and generating the missing types. It is perhaps surprising at first look that this technique can survive the inevitable noise that can be introduced by this imputation. However, we note that this could be cast as a data augmentation technique. Data augmentation is common in the deep learning literature, especially for computer vision, where images may be rotated, scaled, or have noise added to make learning algorithms more robust. Our positive results suggest that the imputed data types do not introduce harmful noise, but this may be sensitive to the size of the central analyzer's initial dataset, and should be explored further in future work.

There are several limitations in this study. First, our method allows model training without patient ID matching at the price of requiring the central analyzer to have some matched data and ignoring some interactions among different data types. In our disease prediction example, the central analyzer had fully connected data from one of the 34 states and the separation did not cause significant performance reduction, likely because the cGAN models trained were able to perform relatively well in the inference and each single data type had already captured a decent amount of information internally. However, in some other applications, more direct inter-data type interaction might be crucial. Second, we demonstrate performance using a single data source due to limited medical data availability. To make a more generalizable conclusion, more data sets will be needed in the experiments. Third, in preliminary work on even rarer diseases (prevalence < 1%), even centralized learning was not sufficient to learn usable classifiers. Rare disease cohort identification is an important potential use case for distributed learning methods, so future work will need to investigate improved representations for learning to classify these diseases so that confederated learning can be tested in this challenging setting. Lastly, we define the diseases using diagnosis code. This approach has two limitations. ICD codes are primarily used for billing purposes and therefore may cause some biases in the labels. In addition, silos like pharmacies and labs do not have access to diagnosis codes, which makes it impossible to train models using only pharmacy or lab datasets.

We anticipate that this confederated approach can be extended to more degrees of separation. Other types of separation, such as separation by temporality, separation by insurance plan, separation by healthcare provider can all potentially be explored using a confederated learning strategy. In the next step, we plan to explore meta-learning based confederated learning methods which potentially enable us to conduct confederated learning without need of a central analyzer with fully linked data. In addition, we would like to use more realistic silos separated by all three dimensions. Examples of such silos include but are not limited to healthcare providers with different speciality, different laboratory test providers and pharmacies.

**Funding:** This work was supported by CVS Health, by cooperative agreements U01TR002623 from the National Center for Advancing Translational Sciences (NCATS)/National Institutes of Health, and U01HL121518 from the National Heart, Lung, And Blood Institute/National Institutes of Health.

**Competing interest:** Kathe Fox is a member of a member of Aetna's Analytics. Other authors declare that there are no competing interests

**Author Contribution:** D.L. initiated the project, developed the concepts, developed the method, conducted data processing, designed the experiments and wrote the manuscript. T.M. contributed to development of the method, design the experiments and review of the manuscript. K.F. contributed to design the experiments and review of the manuscript. G.W. contributed to development of the method, design the experiments and review of the manuscript.